\documentclass[conference]{IEEEtran}
\IEEEoverridecommandlockouts
\usepackage{cite}
\usepackage{amsmath,amssymb,amsfonts}
\usepackage{graphicx}
\usepackage{textcomp}
\usepackage{xcolor}

\usepackage{subcaption}
\usepackage[normalem]{ulem}

\usepackage{xcolor}

\usepackage[utf8]{inputenc}
\usepackage{fourier} 
\usepackage{array}
\usepackage{makecell}

\usepackage{caption}

\def\BibTeX{{\rm B\kern-.05em{\sc i\kern-.025em b}\kern-.08em
    T\kern-.1667em\lower.7ex\hbox{E}\kern-.125emX}}

\begin{document}
\IEEEpubid{\begin{minipage}[t]{\textwidth}\ \\[10pt]
        \raggedright\normalsize{979-8-3503-9341-5/23/\$31.00 \copyright 2023 IEEE}
\end{minipage}} 



\title{Unsupervised Deep Learning Image Verification Method}

\makeatletter
\newcommand{\linebreakand}{
  \end{@IEEEauthorhalign}
  \hfill\mbox{}\par
  \mbox{}\hfill\begin{@IEEEauthorhalign}
}

\author{\IEEEauthorblockN{Enoch Solomon}
\IEEEauthorblockA{\textit{Department of Computer Science} \\
Virginia State University \\
Richmond, Virginia \\
esolomon@vsu.edu
}

\and
\IEEEauthorblockN{Abraham Woubie}
\IEEEauthorblockA{\textit{Silo AI} \\
Helsinki, Finland\\
Abraham.zewoudie@silo.ai}

\and

\IEEEauthorblockN{Eyael Solomon Emiru}
\IEEEauthorblockA{\textit{Department of Information} \\
Engineering and Computer Science \\
University of Trento, Italy \\
eyael.emiru@studenti.unitn.it}
}

\maketitle

\begin{abstract}

Although deep learning are commonly employed for image recognition, usually huge amount of labeled training data is required, which may not always be readily available. This leads to a noticeable performance disparity when compared to state-of-the-art unsupervised face verification techniques. In this work, we propose a method to narrow this gap by leveraging an autoencoder to convert the face image vector into a novel representation. Notably, the autoencoder is trained to reconstruct neighboring face image vectors rather than the original input image vectors. These neighbor face image vectors are chosen through an unsupervised process based on the highest cosine scores with the training face image vectors.  The proposed method achieves a relative improvement of 56\% in terms of EER over the baseline system on Labeled Faces in the Wild (LFW) dataset. This has successfully narrowed down the performance gap between cosine and PLDA scoring systems.

\end{abstract}

\begin{IEEEkeywords}
autoencoder, biometrics, deep learning, face recognition, image recognition
\end{IEEEkeywords}

\section{Introduction} 

Face recognition is a system used for identifying or verifying a person from an image or video \cite{jain2011handbook,solomon2023face,solomon2023fass,solomon2023hdlhc,solomon2023deep,solomon2023autoencoder,woubie2023image,https://doi.org/10.25772/re06-av14}. Deep learning techniques, as a front-end approach, have demonstrated proficiency in acquiring intricate features \cite{liu2015deep}, including bottleneck features (BNF) \cite{deng2014deep}. These features are subsequently integrated into traditional frameworks or employed in the extraction of face image vectors for the face recognition process. Deep learning methods prove effective in compressing face images into embeddings, as exemplified in \cite{snyder2018x} in addition to different speech and security applications \cite{woubie2021federated, woubie2021voice,woubie2021federatedondevice,woubie2021use,cima,scope,ics_sea,memory_safety,fmemory_safety,earic,containers_security}. 

Several works have been proposed to enhance face image verification performance through unsupervised learning techniques \cite{liu2014facial,kurup2019semi,teh2000rate}. For instance, \cite{alphonse2021multi} introduces a vector representation of face images using RBM adaptation. On the backend, \cite{nguyen2010cosine} proposes various imposter selection algorithms aimed at narrowing the performance gap between supervised and unsupervised face recognition methods. They apply DBN adaptation as a backend for face images. These algorithms necessitate training a separate model for each target person, incurring computational costs. However, the results still fall short of the supervised approach, which benefits from actual face image labels.

In this work, we focus on mitigating the necessity for labeled face image data in face verification. Our objective is to narrow the performance disparity between supervised and unsupervised techniques, all without relying on face image labels. Unlike traditional DNN-based classifiers, the training of autoencoders is an unsupervised process that operates independent of labeled data. Our contribution lies in the novel training approach for the autoencoder, specifically designed to account for session variability among face images in the absence of labeled data. Rather than reconstructing the same training face image, we train the autoencoder to reconstruct neighboring face images. Following this training, we extract face image vector embeddings for testing purposes. During experimental trials, we apply cosine scoring to evaluate the face image embedding vectors, which have exhibited a remarkable discriminatory capability. The experimental findings demonstrate a substantial 56\% enhancement in performance over the baseline system that employs the cosine scoring technique, underscoring the efficacy of our proposed autoencoder training methodology.

The rest of the paper is as follows: Section 2 provides an in-depth description of the proposed autoencoder training method and the process of neighbor face image selection. Section 3 describes the experimental configuration and outlines the dataset employed. Section 4 describes the experimental results. Finally, Section 5 gives the conclusion of the proposed work.

\begin{figure}[]
	\centering
		\includegraphics[scale=0.4]{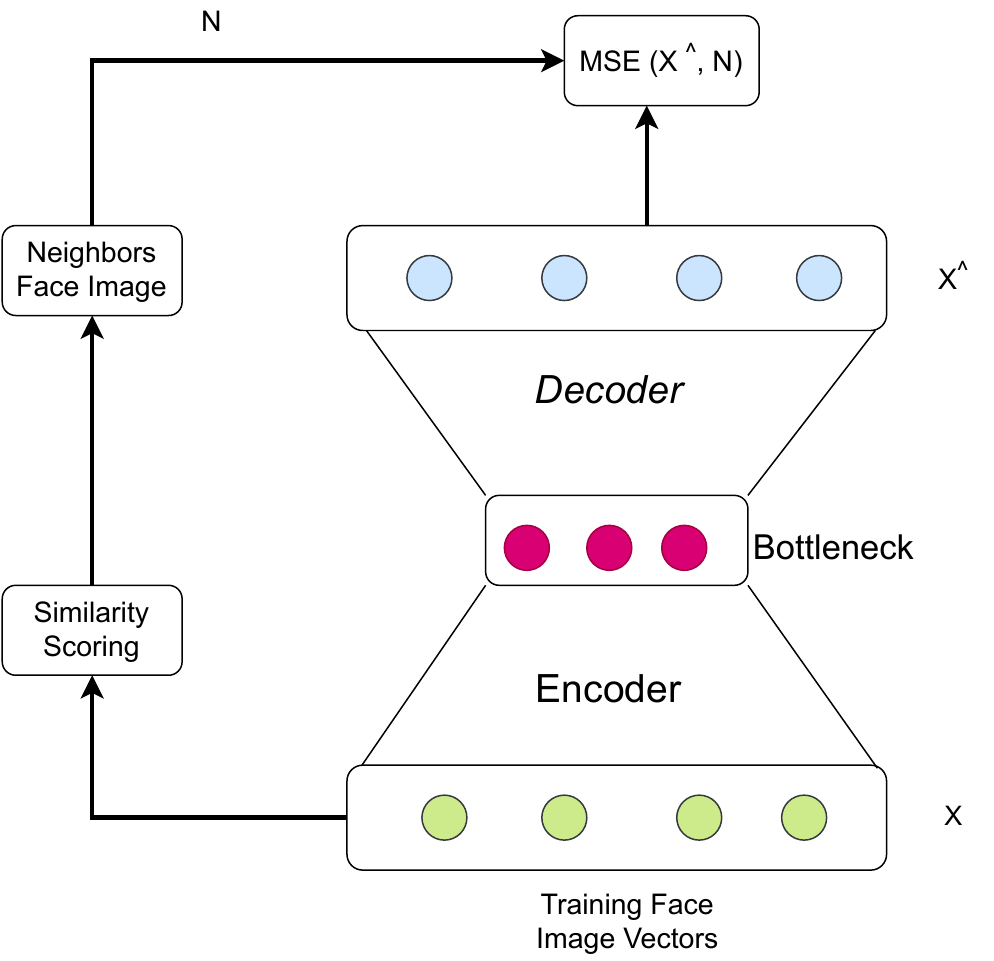}
	\caption{Proposed approach for training the autoencoder.}
	\label{fig:arch2}
\end{figure}

\section{Proposed Method}

In this work, we introduce an unsupervised deep learning method for face verification. The training procedure of the proposed method is outlined as follows: From a pool of unlabeled large-scale face image data, we identify the k nearest neighbor images that closely resemble the specified face image, which are then employed for training. Diverging from the conventional deep neural network training, our method computes the loss function for each input face image k times.   

PLDA scoring technique for face image vectors necessitates labels \cite{muslihah2020texture}, which can be challenging to obtain in practical scenarios. Conversely, the cosine scoring technique operates without the need for labels, albeit at the expense of performance degradation. To bridge the achievement gap between these two scoring methods in the absence of labeled training data, we introduce a novel framework for autoencoder training. This approach is entirely unsupervised, distinguishing it from conventional DNN classifiers and PLDA. Additionally, we advocate training the autoencoder to reconstruct similar face images, rather than exclusively focusing on the same training face image. Through this method, the autoencoder, trained in an unsupervised manner, effectively addresses session variability among face images without relying on labels.

\subsection{Autoencoder training}

As shown in Fig 1, an autoencoder comprises an encoder and a decoder. The encoder functions to compress the input face image vector $x$ into a lower-dimensional space, while the decoder reconstructs it back to its original form. Conventional training involves minimizing the Mean Square Error (MSE) between the input $X$ and the reconstructed $X^{\wedge}$.

In this work, we proposed the training of an autoencoder to  minimize the loss function $M S E\left(X^{\wedge}, N\right)$, illustrated in Figure 1, where $N$ represents a face image vector similar to $X$ and $X^{\wedge}=$ decoder $(\operatorname{encoder}(X))$. We put forth a method for the automatic selection of similar face image vectors. For each training input face image vector $X$, multiple similar face image vectors can be taken into consideration. Thus, each images in the dataset could have different number of similar images. In Section 4, we will conduct a comparative analysis between the results obtained from our proposed training approach and the conventional training method, which involves reconstructing the same training face image vectors. 

\subsection{Selection of similar face image vectors}

All the training face image vectors are assessed against one another using the cosine scoring technique. For each face image vector, a group of similar face image vectors is chosen. A simple method for selecting these similar face image vectors involves setting a threshold on the cosine scores between the training face image vectors. Face image vectors with scores surpassing this threshold are designated as similar face image vectors. It's worth noting that this approach may result in a varying number of similar face image vectors for each training face image vector.

Another approach is to adopt a constant value, denoted as  k, and choose k similar face image vectors for each training face image vector. This method ensures a consistent number of similar face images for every training face image vector, resulting in a balanced training approach.

Figure 2 provides a visual representation of the process for selecting similar face image vectors when a constant $k$ number of neighbors is considered. Let $X_i$ be a training face image vector, where $i=(1, \ldots, n)$ and $n$ represents the total number of training face image vectors. Initially, we compute cosine scores for all training face image vectors relative to one another. Subsequently, we pick the top $k$ face image vectors with the highest scores to serve as similar face image vectors for each $X_i$.  These selected similar face image vectors are denoted as $N_{i j}$, where $N_{i j}$ is the $j^{\text {th }}$ similar of $i^{\text {th }}$ training face image vectors.

Thus, we have a total of $n \times(k-1)$ training data, which automatically generated on the fly which address the issue of having large training data. We run extensive experiment to determine the the values of threshold and $k$. Thus, the autoencoder is able to learn information about the intrinsic variability of face image vectors, without necessarily using actual face image labels.

Consequently, we accumulate a total of $n \times(k-1)$ samples for the autoencoder training process. The specific values for the threshold and $k$ are established through experimental investigation, and a detailed discussion on these parameters will be provided in Section 4. Through this approach, the autoencoder effectively assimilates information regarding the session variability inherent in face image vectors, all without the dependency on actual face image labels.

\subsection{Autoencoder vector extraction}

After the autoencoder is trained with the chosen neighbor face image vectors, we proceed to convert the testing face image vectors into a fresh face image vector representation, utilizing the autoencoder, as illustrated in Figure 1. At this stage, we extract the target face image vectors from the output of the autoencoder. In the experimental trials, these extracted vectors have demonstrated an enhanced discriminatory capability for face image vectors, all achieved without reliance on face image labels. Leveraging these extracted vectors, we carry out the experimental trials using the cosine scoring technique.

\begin{figure}[]
	\centering
		\includegraphics[scale=0.4]{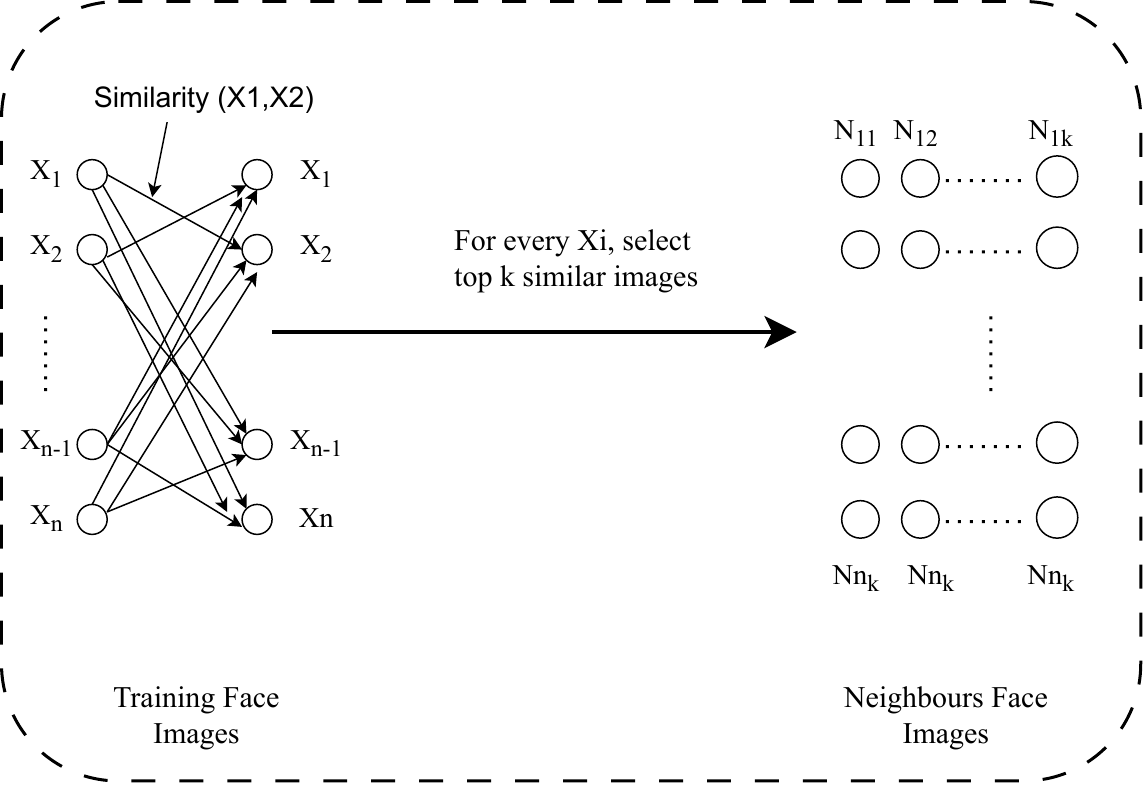}
	\caption{Proposed method of selecting the neighbor face image.}
	\label{fig:arch}
\end{figure}

\section{Experimental Setup and Dataset}

\subsection{Experimental Setup}

For the training of our model, we utilized the Keras deep learning library \cite{chollet2015keras}. The architecture of the autoencoder employed in this work comprises three hidden layers. This design maintains symmetry between the encoder and decoder sections. Specifically, both the first and third hidden layers contain an identical number of neurons, as illustrated in Fig. 2. Each of the encoder input and decoder output layers consists of 112 by 112 neurons. In the second layer of both the encoder and decoder, there are 800 neurons. The encoder's output layer is composed of 300 neurons, determining the size of the resulting embedding vector. Within the DNN, the dimension of the face image embedding layer was fixed at 400, while the classification layer encompassed 1000 neurons.

The training of the autoencoder was conducted over 400 epochs or until the error stops decreasing and showed no further decrease. In all layers of the autoencoder, we applied the Rectified Linear Unit (ReLU) activation function, with the exception of the final layer, which utilized a linear activation function. The learning rate was set to 0.03, incorporating a decay rate of 0.0002. We implemented a logarithmically decaying learning rate, spanning from $10^{-2}$ to $10^{-8}$. The batch size was configured at 100.

\subsection{Datasets}

The CelebA dataset, introduced by Liu et al. \cite{liu2015deep}, encompasses a vast collection of over 200,000 images featuring 10,177 celebrities. This dataset incorporates diverse elements such as pose variations and background clutter, providing a comprehensive representation of real-world scenarios. To facilitate model training and evaluation, the dataset is partitioned into distinct sets for training, validation, and testing purposes. In this study, we leveraged the training and test segments of the CelebA dataset as our unlabeled dataset for training the autoencoder. In contrast, supervised training was conducted using the validation portion of the CelebA dataset.

The Labeled Faces in the Wild dataset (LFW) \cite{huang2008labeled} comprises a collection of 13,233 images featuring 5,749 distinct individuals. For testing purposes, this database is randomly and uniformly divided into ten subsets. Within each subset, 300 pairs of matched images (depicting the same person) and 300 pairs of mismatched images (depicting different persons) are randomly selected. In essence, this testing protocol employs a total of 3,000 matched pairs and 3,000 mismatched pairs \cite{huang2008labeled}.

\section{Experimental Results}
\subsection{Comparison with different threshold values}

\begin{table}[]
\caption{The Equal Error Rate (EER) was computed for both the proposed face image vectors and the baseline, assessed across various threshold values when employing cosine scoring.}
\label{table:1}
\centering
\begin{tabular}{lllll}
\hline
\thead{Approach} & \thead{threshold} &   \thead{EER(\%)} \\ \hline
[1] Baseline & No & 16.84
\\ \hline
[2] Proposed & 0.7 & 15.30
\\ \hline
[3] Proposed & 0.6 & 15.04
\\ \hline
[4] Proposed & 0.5 & 13.82
\\ \hline
[5] Proposed & 0.4 & 11.23
\\ \hline
[6] Proposed & 0.3 & 9.61
\\ \hline
[7] Proposed & 0.2 & 8.13
\\ \hline
[8] Proposed & 0.1 & 8.24
\\ \hline
Fusion of [1] and [7] & combined & 7.12
\\ \hline

\end{tabular}
\end{table}

In Tables I and II, approach refers to the method where face image vectors are extracted using a conventionally trained autoencoder. Table I presents a performance comparison between our proposed method and the baseline, where a threshold is set to select a specific number of neighbor face image vectors. All vectors are then evaluated using the cosine scoring technique. We experimented with different threshold values to fine-tune the system and achieve the best results. The table clearly illustrates that our proposed method outperforms the baseline system. As we decrease the threshold value, the system's performance improves. The optimal Equal Error Rate (EER) of 8.24\% was achieved with a threshold set to 0.2. This represents a 51\% relative improvement over the baseline method. Consequently, this bridges the performance gap between cosine and Probabilistic Linear Discriminant Analysis (PLDA) scoring techniques.

By performing a score-level fusion of the baseline and our proposed method, using a threshold of 0.2, we achieved a further enhancement in performance, as evidenced by a reduced Equal Error Rate (EER) of 7.12\%. The optimal weights for the fusion were determined empirically, resulting in values of 0.55 and 0.45 for the baseline and our proposed method, respectively. This fusion strategy effectively leverages the strengths of both approaches to yield superior results.

\subsection{Comparison with different k values}

\begin{table}[]
\caption{The Equal Error Rate (EER) was computed for both the proposed face image vector and the baseline method across different values of k all evaluated using the cosine scoring technique.}
\label{table:2}
\centering
\begin{tabular}{lllll}
\hline
\thead{Approach} & \thead{k} &   \thead{EER(\%)} \\ \hline
[1] Baseline & No & 18.54
\\ \hline
[2] Proposed & 1 & 15.21
\\ \hline
[3] Proposed & 2 & 13.13
\\ \hline
[4] Proposed & 3 & 12.71
\\ \hline
[5] Proposed & 4 & 10.12
\\ \hline
[6] Proposed & 10 & 8.40
\\ \hline
[7] Proposed & 11 & 8.03
\\ \hline
[8] Proposed & 12 & 8.94
\\ \hline
Fusion of [1] and [11] & combined & 7.01
\\ \hline

\end{tabular}
\end{table}

In Table II, we present a performance comparison between our proposed method and the baseline, utilizing a constant value of k for the selection of similar face image vectors. We conducted experiments with various values of 
k to optimize the system for achieving the best results. The table illustrates that employing a fixed value of 
k has demonstrated improvement compared to both the threshold-based approach and the baseline. Beginning with 
k set to 1, an Equal Error Rate (EER) of 15.21\% was attained. This signifies a 17\% relative improvement over the baseline system by considering only the nearest neighbor for each image vector. As we increment the value of 
k, the system's performance continues to improve. The most favorable EER of 8.03\% was achieved when 
k was set to 11, leading to a remarkable 56\% relative improvement over the baseline method. This achievement effectively bridges the performance gap between cosine and PLDA scoring techniques. Moreover, this approach enables a balanced training regimen, ultimately enhancing the system's performance. It is worth noting that a further increase in the value of k may incorporate very distant neighbors, potentially degrading the system's performance.

By combining the scores from the baseline and our proposed method, both utilizing 
k set to 11, we achieved a further improvement in performance, specifically in terms of the Equal Error Rate (EER). This fusion approach yielded an EER of 7.01\%, which is in close proximity to the results obtained using Probabilistic Linear Discriminant Analysis (PLDA) with actual face image labels. The optimal weights for the fusion were fine-tuned through experimental iteration and were ultimately set to 0.49 and 0.51 for the baseline and our proposed method, respectively. This fusion strategy effectively leverages the strengths of both approaches to yield highly competitive results.

\subsection{Comparison with  score level fusion}

\begin{table}[]
\caption{The Equal Error Rate (EER) was computed for the score-level fusion between Probabilistic Linear Discriminant Analysis (PLDA) and cosine scoring, using both the proposed face image vectors with 
k set to 15, and the baseline method.}
\label{table:3}
\centering
\begin{tabular}{lllll}
\hline
\thead{Approach} & \thead{Scoring} &   \thead{EER(\%)} \\ \hline
[1] Baseline & PLDA & 6.71
\\ \hline
[2] Proposed (k = 11) & Cosine & 8.03
\\ \hline

Fusion of [1] and [2] & combined & 6.23
\\ \hline

\end{tabular}
\end{table}

As depicted in Table III, conducting a score-level fusion between Probabilistic Linear Discriminant Analysis (PLDA) and cosine scoring, utilizing the proposed method with 
k set to 11 (yielding the best results), leads to an Equal Error Rate (EER) of 6.23\%. This showcases an impressive relative improvement of nearly 7\% over the PLDA approach alone. The optimal fusion weights were determined to be 0.04 for the baseline and 0.96 for the proposed method, respectively. This fusion strategy effectively capitalizes on the strengths of both approaches, resulting in a highly competitive performance.

\subsection{Comparison with the state of the art on LFW dataset}

\begin{table}[]
\caption{Comparison of the proposed method result with some of state-of-the-art methods on LFW testing dataset.}
\begin{tabular}{lllll}
\hline
\thead{Model} & \thead{Training \\ size} &  \thead{Labeled/ \\ Unlabeled} &  \thead{Testing \\ size} & \thead{Accuracy(\%)} \\ \hline

UniformFace      & 6.1M                  & Labeled    & 6K                                    &  99.80 \cite{duan2019uniformface}                      \\ \hline

ArcFace               &  5.8M                           & Labeled   & 6K                                              & 99.82 \cite{deng2019arcface}                        \\ \hline

GroupFace  & 5.8M & labeled & 6K   & 99.85 \cite{kim2020groupface}                                    \\ \hline

CosFace               &  5M                           & Labeled    & 6K                                             & 99.73 \cite{wang2018cosface}                        \\ \hline

Marginal Loss               &  4M                           & Labeled   & 6K                                             & 99.48 \cite{deng2017marginal}                        \\ \hline

CurricularFace               &  3.8M                           & Labeled   & 6K                                             & 99.80 \cite{huang2020curricularface}                        \\ \hline

RegularFace               &  3.1M                           & Labeled    & 6K                                            & 99.61 \cite{zhao2019regularface}                        \\ \hline

AFRN               &  3.1M                           & Labeled      & 6K                                          & 99.85 \cite{kang2019attentional}                        \\ \hline

 Stream Loss               &  1.5M                           & Labeled     & 6K                                          & 98.97 \cite{rashedi2019stream}                        \\ \hline

MDCNN               &  1M                           & Labeled    & 6K                                            & 99.38 \cite{huang2022face}                        \\ \hline
  
  ULNet               &  1M                          & Labeled     & 6K                                           & 99.70 \cite{boragule2022learning}                        \\
 
 \hline
Ben Face               &  0.5M                          & Labeled    & 6K                                            & 99.20 \cite{ben2021face     }                   \\
 
 \hline

 F$^2$C               &  5.8M                          & Labeled   & 6K                                             & 99.83 \cite{wang2022efficient}                       \\
 \hline
PCCycleGAN               &  0.5M                          & Unlabeled    & 6K                                            & 99.52 \cite{liu2021unsupervised}                        \\
 
 \hline
CAPG GAN               &  1M                          & Unlabeled    & 6K                                            & 99.37 \cite{hu2018pose}                        \\

 \hline
 UFace               &  200K                          & Unlabeled    & 6K                                            & 99.40 \cite{solomon2022uface}                        \\

 \hline
 
\textbf{Proposed}      & \textbf{200K}                             & \textbf{Unlabeled}    & \textbf{6K}                                            & \textbf{99.53}              \\ \hline

\end{tabular}
\end{table}

Despite the slightly higher accuracy of methods like ArcFace, GroupFace, Marginal Loss, and CosFace, it's noteworthy that our proposed method achieves competitive results with a significantly smaller training dataset (around 200K images). In contrast, many state-of-the-art methods rely on much larger training sets, often in the millions of images. This demonstrates the efficiency and effectiveness of our approach, particularly in scenarios where acquiring a massive dataset may not be feasible or practical.

The experimental result of the proposed method become particularly notable when compared to state-of-the-art approaches. For instance, ArcFace, which attains 99.82\% accuracy, utilized a substantially larger dataset of 5.8 million labeled images. In contrast, our method achieved comparable performance with significantly fewer training images (around 200K). This underscores the efficiency and effectiveness of our approach, especially in scenarios where obtaining a massive dataset is challenging.

\section{Conclusion}

The proposed work introduces a novel unsupervised deep learning method for face verification. The training process involves selecting 
k images from unlabeled data that are most similar to a specified face image, and using them for training. Unlike standard deep neural network training, our proposed method calculates the loss function for each input image 
k times. The primary goal is to bridge the performance gap between cosine and PLDA scoring techniques when labeled training data is unavailable. This is achieved by training an autoencoder to reconstruct a set of neighbor face image vectors, instead of reconstructing the same training face image vectors. These neighbor face image vectors are selected based on their cosine scores in relation to the training face image vector. After training, face image vectors are extracted for test face images as the output of the autoencoder. In experimental trials, face image vectors are scored using cosine scoring. The evaluation conducted on the LFW dataset demonstrates that our proposed method achieves a relative improvement of 56\% in terms of Equal Error Rate (EER) over the baseline system. This successfully narrows the performance gap between cosine and PLDA scoring systems.

\end{document}